\pdfoutput=1

\documentclass[11pt]{article}

\usepackage{acl}

\usepackage{times}
\usepackage{latexsym}

\usepackage[T1]{fontenc}
\usepackage{placeins}
\usepackage[utf8]{inputenc}

\usepackage{microtype}

\usepackage{hyperref}
\usepackage{url}
\usepackage{subcaption}

\usepackage{booktabs}
\usepackage{amsmath}
\usepackage{booktabs}

\definecolor{highlight}{RGB}{44, 123, 182}
\definecolor{accent}{RGB}{255, 127, 14}

\usepackage{amsfonts}

\usepackage{xifthen} 
\usepackage{calc}    

\usepackage{inconsolata}

\usepackage[affil-it]{authblk}
\usepackage{marvosym}

\usepackage{graphicx}

\title{Working Memory Constraints Scaffold Learning in Transformers \\ under Data Scarcity}

\author[$\triangle$,$\infty$]{Pranava Madhyastha}
\author[$\oplus$]{Dagmar Adamcová}
\affil[$\triangle$]{City, University of London}
\affil[$\infty$]{The Alan Turing Institute}
\affil[$\oplus$]{Grounded Machines}
\affil[\Email]{pranava.madhyastha@city.ac.uk, dagmar@groundedmachines.com}

\begin{document}
\maketitle
\begin{abstract}

We investigate the integration of human-like working memory constraints into the Transformer architecture and implement several cognitively inspired attention variants, including fixed-width windows based and temporal decay based attention mechanisms. Our modified GPT-2 models are trained from scratch on developmentally plausible datasets (10M and 100M words). Performance is evaluated on grammatical judgment tasks (BLiMP) and alignment with human reading time data. Our results indicate that these cognitively-inspired constraints, particularly fixed-width attention, can significantly improve grammatical accuracy especially when training data is scarce. These constrained models also tend to show a stronger alignment with human processing metrics. The findings suggest that such constraints may serve as a beneficial inductive bias, guiding models towards more robust linguistic representations, especially in data-limited settings.
\end{abstract}

\section{Introduction}

\label{sec:introduction}
The dominant self-attention mechanism in Transformer-based models \citep{Vaswani+2017}  diverges profoundly from established cognitive theories of human language processing. A key area of divergence lies in how information is accessed and maintained over time. Human comprehenders rely on working memory, a short-term mental buffer that temporarily stores and manipulates information during language processing. This memory system is understood to be capacity-limited \citep{miller1956magical, cowan2001magical}, subject to temporal decay \citep{baddeley2000episodic}, and influenced by serial position effects like primacy and recency \citep{glanzer1966two}. In stark contrast, standard Transformer self-attention allows near-uniform access to all tokens within a potentially very large context window, lacking inherent architectural biases that reflect these fundamental human cognitive constraints. While language models like GPT-2 have proven useful in cognitive modelling, correlating well with measures such as reading times and neural activity \citep{goodkind-bicknell-2018-predictive,wilcox-etal-2020-structural,madhyastha2023words}, this success may occur despite, rather than because of, architectural alignment with human processing limitations.

This discrepancy motivates the central research question of our work: can integrating architectural constraints inspired by human working memory (henceforth refered as WM) produce models that not only exhibit more human-like linguistic behaviour but also learn more efficiently? To this end, our contribution is threefold. First, we implement and systematically compare four distinct, cognitively-motivated attention mechanisms: (1) a strict, fixed-width attention window to model capacity limits; (2) exponential and (3) logistic decay mechanisms to model recency bias; and (4) a primacy-recency model to capture serial position effects. Second, we train these models entirely from scratch on developmentally plausible datasets of 10 million and 100 million words, allowing these constraints to shape the learning process from its inception. Third, we conduct a multi-faceted evaluation, assessing both grammatical competence on the BLiMP benchmark \citep{warstadt2020blimp} and alignment with human processing data and the nature of the internal representations learned by the models.

Our results show that imposing attention constraints, particularly
restrictive fixed-width attention windows, serves as a powerful
inductive bias that yields significant benefits. In low-data settings,
constrained models substantially outperform the standard GPT-2
baseline in grammatical accuracy, confirming that such biases are
highly effective when data is scarce. We also see that these models
produce surprisal values that are markedly more predictive of human
processing data, suggesting a closer alignment with the cognitive
dynamics of comprehension. While this performance gap narrows with
more training data, psychometric alignment in fact degrades for all
models at the larger scale, consistent with a growing body of evidence
that the language-modelling objective and human comprehension are not
asymptotically convergent \citep[see for e.g.,][]{oh2023does, shain2024large,
de2024cloze}. Attention visualisations and structural probing further clarify why these constraints help. Limiting the context forces the model to develop specialised, linguistically interpretable attention heads and a more explicit internal encoding of syntactic structure. The unconstrained baseline shows no comparable specialisation, with attention patterns that remain diffuse across layers.

\section{Background}

\label{sec:background}
In psycholinguistics, verbal WM constraints are well-documented experimentally. For example, long or structurally
complex sentences have been shown to impose a burden on memory resources
\citep{caplan2013memory, just1992capacity}, and comprehenders
appear to manage this load through incremental processing that
prioritises local dependencies, with representations of earlier
material becoming progressively less accessible
\citep[see][\textit{inter alia}]{levy2008expectation, hahn2022resource,
futrell2020lossy, gibson1998linguistic}. A strong claim,
advanced by \citet{christiansen2016now} and others
\citep[e.g.,][]{kirby1999function}, is that the structure of human
language itself reflects these constraints: features ranging from
preferred word orders to limits on centre-embedding may be
adaptations to a working memory bottleneck. Experimental support in this regard comes from typological evidence that dependency lengths are systematically minimised across languages \citep{futrell2015large}.

Recent work in computational psycholinguistics has begun to probe how these constraints might be reflected in language models, often within the developmentally plausible data scales (10–100M words) emphasised by the BabyLM Challenge \citep{warstadt-etal-2023-findings}, which enables investigation of cognitive constraints in regimes more comparable to human language acquisition. \citet{ryu-lewis-2021-accounting} show that pre-trained GPT-2 implicitly captures similarity-based interference effects, and \citet{timkey2023language} advocate simpler architectures grounded in cue-based retrieval theories \citep{van2003distinguishing}. A more direct line of work modifies attention itself: \citet{de-varda-marelli-2024-locally} apply an exponential decay bias to a pre-trained GPT-2 model and report improved reading-time prediction; \citet{kuribayashi-etal-2022-context} show that truncating context at inference improves cognitive alignment of surprisal; \citet{clark-etal-2025-linear} train with ALiBi \citep{press2021train}, a linear bias that softly downweights distant tokens. \citet{janik2023aspects} reports that standard Transformers exhibit weak primacy and recency effects emergently, though these are scale-sensitive and appear to be statistical artefacts of training.

These interventions on attention share a common motivation but each leaves the constraint operating at the periphery of the model. Post-hoc decay biases \citep{de-varda-marelli-2024-locally} and inference-time context truncation \citep{kuribayashi-etal-2022-context} modify the output of an already-trained system whose representations were shaped without any such constraint. Soft positional biases such as ALiBi \citep{clark-etal-2025-linear, press2021train} are present during training but only discourage rather than explicitly prevent long-range attention, which may allow the model to bypass the constraint when useful. Emergent effects in unconstrained models \citep{janik2023aspects} are scale-sensitive and fragile. 

In light of these limitations, we integrate working memory constraints directly into the GPT-2 architecture from the outset of training. This contrasts with approaches based on implicit learning or post-hoc modification, which leave the underlying representations unconstrained. Specifically, we investigate how imposing constraints inspired by prominent findings in human working memory research (including capacity limitations, temporal decay, and serial position effects) impacts model performance and language processing.

\section{Methods}
We implement four distinct attention mechanisms directly into the GPT-2 architecture to simulate human-like WM constraints. Each of these mechanisms is designed to model specific features of human WM. 

\subsection{Fixed Window Attention}

A hallmark feature of human working memory is its limited storage capacity. While classic accounts attempted to quantify this capacity as a discrete number of items or ``chunks", arriving at numbers between four and nine \citep{miller1956magical, cowan2001magical}, modern theories however have moved toward more dynamic models. These accounts posit that capacity is not a fixed unit count but is instead a skill that emerges from language comprehension and production processes, viewing verbal WM as the activated portion of linguistic long-term memory. This perspective suggests that the ability to maintain and order verbal information is intrinsically linked to an individual's language experience and proficiency, rather than relying on separate, discrete storage buffers \citep{schwering2020verbal, macdonald2016speak, buchsbaum2019sensorimotor}. Nevertheless, we find it useful to simulate this very restrictive capacity constraint as a discrete cut-off in order to isolate local dependencies. We do this by implementing a fixed-width attention window.

In this approach, the attention calculation for each token is restricted to a \emph{fixed-size window} of a set of fixed preceding tokens. For a token at position $i$, attention is only computed over tokens in the range $[\max(0, i-W+1), i]$, where $W$ is the fixed window size.  This is implemented using an attention mask $M_{window}$ that prevents access to tokens outside the window:
\begin{equation}
\resizebox{\columnwidth}{!}{
$
M_{window}^{(i, j)} =
\begin{cases}
    0, & \text{if } \max(0, i-W+1) \leq j \leq i \\
    -\infty, & \text{otherwise}
\end{cases}
$
}
\label{eq:fixed_window_mask}
\end{equation}

The attention weights are then calculated as:
$a'_{ij} = a_{ij} + M_{window}^{(i, j)}$, 
where $a_{ij}$ are the original attention weights. 
This sets the attention weights for tokens outside the window to $-\infty$ (and thus zero after \texttt{softmax} normalisation), while tokens inside the window retain their original weights. The mechanism imposes a hard capacity limit at each layer, forcing the model to operate within a strictly local context.
The selection of fixed window sizes for our models is directly motivated by influential findings in cognitive science concerning the capacity limits of human WM. A window size of $k=4$ is informed by contemporary research, such as that of \citet{cowan2001magical}, who hypothesised that short-term memory has a capacity of approximately four `chunks' of information. The window sizes of $k\in{5,7,9}$ are derived from \citeposs{miller1956magical} seminal observation concerning ``the magical number seven, plus or minus two'', which represents the classic estimate for the number of items an individual can hold in immediate memory. In the context of our experiments, we treat a token as a fundamental information chunk. 

In the context of NLP, a form of fixed window attention is a core component of models designed for long documents, such as Longformer~\citep{beltagy2020longformer} and Block-Sparse Transformer~\citep{child2019generating}. While these models are motivated by efficiency, our fixed window attention is motivated by mirroring the limited capacity of human WM, suggesting a cognitive basis for such architectural choices in NLP.

\subsection{Primacy-Recency Attention}

Human memory recall has been hypothesised to exhibit primacy and recency effects~\citep[][\emph{inter alia}]{glanzer1966two, morrison2014primacy}. Items presented at the beginning (primacy) and end (recency) of a list are typically better recalled than items in the middle. In the context of language processing, this suggests that initial and final parts of a sequence might hold disproportionate importance in shaping the overall representation. We incorporate constraints of this kind through a primacy-recency attention mechanism. This mechanism adds a position-dependent bias to the attention weights that emphasise both the initial and final tokens in the sequence. It learns two parameters, $w_{\text{primacy}}$ and $w_{\text{recency}}$, which are initialized to 0.5 during training. We calculate primacy weights $p_i$ and recency weights $r_i$ for each position $i$ in a sequence of length $L$:

\begin{align}
p_i &= \frac{e^{-i/L}}{\sum_{j=0}^{L-1} e^{-j/L}} \label{eq:primacy_weights} \\
r_i &= \frac{e^{-(L-1-i)/L}}{\sum_{j=0}^{L-1} e^{-(L-1-j)/L}} \label{eq:recency_weights}
\end{align}
where $i$ is the position index (starting from 0). %
Primacy weights decay exponentially from the beginning of the sequence, while recency weights decay exponentially from the end. Both sets of weights are normalised to sum to one. The final bias $b_i$ for each position is a weighted combination of primacy and recency weights: $b_i = w_{primacy} \cdot p_i + w_{recency} \cdot r_i$, 
where $w_{primacy}$ and $w_{recency}$ are learnable weights that control the relative contribution of primacy and recency biases. These biases are then added to the attention weights as:
$
a'_{ij} = a_{ij} + b_j.
$
We note here that the bias $b_j$ is added based on the \emph{key} position $j$ (i.e., the position of the token being attended to in the attention mechanism). Our intention with this position-dependent bias is to encourage the model to attend more strongly to tokens at the beginning and end of the sequence, reflecting primacy and recency effects from psycholinguistic theories. We also separately run an ablation with exclusively primacy and recency based attention to understand the impact of each of these mechanisms. While less directly related to computational efficiency in long sequence processing, the primacy-recency attention mechanism aligns with the broader trend in NLP towards incorporating positional information in more sophisticated ways than simple positional embeddings. For instance, relative positional embeddings~\citep{shaw2018self} and complex positional encodings~\citep{su2021roformer} aim to capture richer positional relationships. In some way, this attention modification helps focus on more positional information to emphasize the structural importance of sequence beginnings and endings.

\subsection{Exponential Decay Attention}
Inspired by recent psycholinguistic theories that highlight the interplay between linguistic expectations and WM constraints in human language processing~\cite[see][\emph{inter alia}]{smith2013effect,gibson1998linguistic,hahn2022resource}, we also consider an exponential decay attention mechanism.
This modification is directly motivated by the work of~\citet{de-varda-marelli-2024-locally}, who propose biasing Transformer models to prioritize local linguistic context, simulating a lossy representation of distant contextual information in human sentence processing. Here, the exponential decay attention mechanism modulates the standard attention weights by incorporating a decay factor that diminishes the influence of tokens based on their temporal distance.  The modified attention weight $a'_{ij}$ between token $i$ and token $j$ is calculated as:

\begin{equation}
a'_{ij} = (1 - \alpha) a_{ij} + \alpha e^{-|i-j| \cdot \lambda}
\label{eq:exponential_decay_attention_devarda}
\end{equation}
where $a_{ij}$ represents the original dot-product attention weight, $\lambda$ is the decay rate, and $\alpha$ is a mixing parameter. The exponential term $e^{-|i-j| \cdot \lambda}$ introduces a bias favouring attention to closer tokens, effectively implementing a recency effect by exponentially reducing the contribution of more distant tokens. Following \citet{de-varda-marelli-2024-locally}, we adopt the hyperparameters $\lambda=82.86$ (corresponding to \texttt{decay\_rate} in our implementation) and $\alpha=0.37$. These values were identified as optimal in their grid search using GPT-2-small on the Provo corpus \citep{luke2018provo}. \citet{de-varda-marelli-2024-locally} demonstrated that applying a post-hoc exponentially decaying attention bias to a pretrained GPT-2 model improved its correlation with human reading times. While these results are useful, this approach modifies an already developed system rather than allowing constraints to shape the learning process from the outset. To remedy this, our methodology involves training the customised GPT-2 model from scratch with the exponential decay attention mechanism inherently integrated into its architecture. This approach will allow the model learns to process language under the constraint of locality-biased attention from the outset. We hypothesise that training with this constraint from the beginning may lead to a more concordant and effective integration of the psycholinguistic principle, as the model architecture is aligned with the intended processing mechanism throughout the learning process, rather than having the bias imposed after the model has already learned with a different attention paradigm. %
This approach has clear parallels within the field of NLP, where locality-sensitive attention is explored for efficiency reasons when handling long sequences. This is evident in models such as Performer~\citep{choromanski2020rethinking} and in Transformers that use linear biases~\citep{katharopoulos2020transformers, press2021train}.

\subsection{Logistic Decay Attention}

While exponential decay offers a straightforward way to model temporal forgetting, its immediate and sharp decline may not accurately reflect how WM handles recent verbal material. This is particularly relevant when considering the recency effect, where the most recent information remains highly accessible for a short period before its recall probability diminishes \citep{glanzer1966two}. To reconcile this with the concept of a limited memory span, we introduce a second temporal decay mechanism: logistic decay. Unlike an exponential bias, which applies a consistent rate of forgetting from the outset, a logistic function imposes a non-linear S-shaped curve. We hypothesise that this curve can more realistically represent the dynamics of human WM by providing a short period of sustained high accessibility for recent tokens before a rapid drop-off in influence, thereby combining elements of both a discrete capacity cut-off as reflected in our fixed-window attention and a more nuanced decay function.
The logistic decay mechanism modulates attention weights based on the temporal distance between tokens using a psychometric function. For tokens at positions $i$ and $j$, the attention weight modification is computed as:
    \begin{equation}
    w_{ij} = \frac{1}{1 + e^{k \cdot (d_{ij} - m)}}
    \label{eq:logistic_decay_weight}
    \end{equation}
    where $d_{ij} = \max(1, i - j + 1)$ represents the distance between tokens, $k$ is the steepness parameter controlling the sharpness of the decay curve, and $m$ is the midpoint parameter determining the distance at which attention weight equals 0.5. The final attention weights are calculated by multiplicatively combining the original attention scores with the logistic prior:
$ a'_{ij} = a_{ij} \cdot w_{ij}$.
We set $k = 0.4$ and $m = 12.0$ as default parameters, establishing a psychologically motivated attention profile where tokens within approximately 5 positions maintain relatively strong (high) attention weights, while more distant tokens experience rapid attention decay. The logistic decay attention mechanism exhibits several key characteristics that distinguish it from other approaches. First, it maintains relatively stable attention weights for nearby tokens (distance $< m$), followed by a rapid transition to low attention weights for distant tokens (distance $> m$). This creates a more pronounced boundary between accessible and inaccessible memory, aligning with accounts of discrete WM span. Second, unlike fixed window attention which implements a hard cutoff, logistic decay provides a smooth but steep transition, avoiding the potential discontinuities associated with binary attention masking.

\section{Experimental Setup}
\label{sec:experimental_setup}

This section details our experimental setup. We introduce the set of models evaluated, the datasets employed for training, and the overall framework for our analysis, which focuses on training language models from scratch with modified attention mechanisms.
Our experiments are conducted on the GPT-2-small architecture \citep{radford2019language}. Our core models are based on the standard GPT-2 configuration, but incorporate custom attention mechanisms implemented by modifying the \texttt{GPT2Attention} module. We also train two baseline models using the default GPT-2 configuration provided by the Hugging Face \texttt{transformers} library \citep{wolf2019-etal-2020-transformers}. %

\subsection{(Pre-)Training Corpora}
\label{subsec:training_data}

Our primary interest lies in evaluating language models under conditions that are more cognitively plausible in terms of data scale than typical large language model pretraining. Therefore, we utilize the BabyLM dataset \citep{warstadt-etal-2023-findings}. The BabyLM Challenge itself drew inspiration from the scale of data available during human language acquisition. 
Specifically, we use the training portions of "Strict-Small" (10 million words) and "Strict" (100 million words) training subsets provided by the BabyLM Challenge \citep{warstadt-etal-2023-findings}. These datasets comprise text from sources considered potentially relevant to child language exposure, including Simple English Wikipedia, children's books from Project Gutenberg, CHILDES transcripts, the British National Corpus, OpenSubtitles, and the Switchboard Dialog Act Corpus. We note that all of our models also use GPT2Tokenizers which are trained specifically on 10 million and 100 million words based corpora separately. 
\subsection{Training Configuration}
\label{subsec:training_config}

For all models, we used the GPT-2 small architecture as a base. Training was performed using the AdamW optimizer with a learning rate of $5e^{-5}$, a batch size of $64$, and a weight decay of $0.01$. Models were trained for 5 epochs with a batch size of $50$. Gradient clipping was applied with a maximum norm of 1.0. These settings were kept consistent across all model variants to ensure a fair comparison. These hyperparameters are similar to the range of  empirical setups common in \citet{warstadt-etal-2023-findings}.

\subsection{Tasks}
\label{subsec:tasks}

We evaluate the trained models on two distinct tasks designed to probe different aspects of their linguistic capabilities and cognitive plausibility.

\paragraph{BLiMP} The first task assesses the sensitivity of the models to English grammatical structure using the Benchmark of Linguistic Minimal Pairs (BLiMP) \citep{warstadt2020blimp}. BLiMP consists of numerous sub-tasks, each targeting a specific linguistic phenomenon. Every example in BLiMP presents a "minimal pair": one sentence that is grammatically acceptable and another that is unacceptable, with the two differing only minimally (often by a single word or morpheme). A language model is considered correct on a given minimal pair if it assigns a \emph{higher probability score} (hence low surprisal) to the acceptable sentence compared to the unacceptable one. Success across BLiMP tasks indicates that the model has learned representations consistent with fine-grained grammatical distinctions in English.

\paragraph{Psychometric Benchmark} The psychometric data we use are drawn from established experimental paradigms (detailed in \citealt{de2024cloze}) with measurements averaged across several neural and behavioural indices of cognitive processing on a set of 1725 sentences. This includes a) Eye-Tracking Data with measures such as First Fixation Duration, which reflects early lexical access, and later-stage integrative measures like Gaze Duration, Go-Past Time, and Right-Bounded Time; b) Self-Paced Reading Time which is a controlled measure of reading speed, influenced by a range of semantic and syntactic factors; c) Event-Related Potentials which are neural signals that provide a fine-grained temporal view of language processing. These include the N400 component, which is modulated by meaning processing; the P600, indicative of syntactic reanalysis and integration; and various Left Anterior Negativity components (LAN and ELAN) associated with phrase structure building. 

\section{Results}

\begin{figure}
    \centering
    \includegraphics[width=\linewidth]{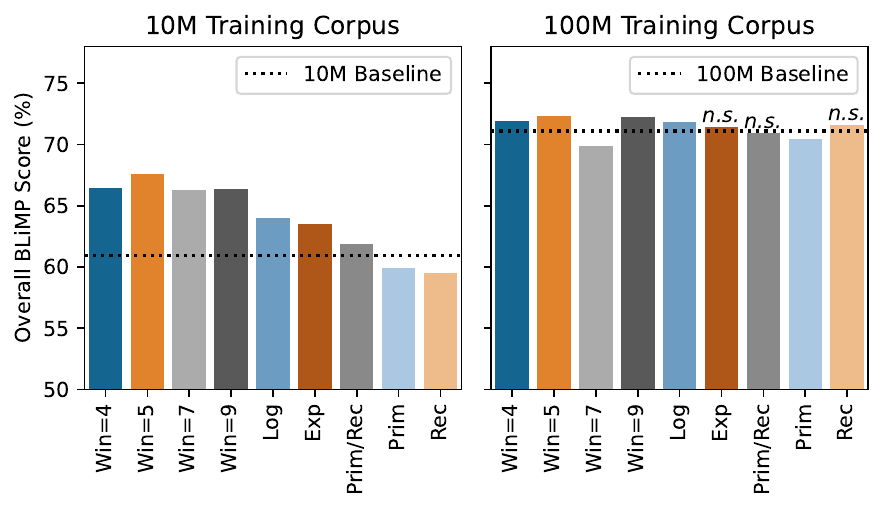}
    \caption{Performance comparison of GPT-2 self-attention modifications on BLiMP tasks for 10M (left) and 100M (right) parameter models. 
Dashed lines indicate the performance of the baseline GPT-2 model (without any attention modifications) trained from scratch on the corresponding dataset size. All models show statistically significant differences from baseline (p < 0.001) except where marked as not significant (n.s.).}
    \label{fig:blimpbar}
\end{figure}

\subsection{Overall observations on grammaticality}
\label{sec:overallob}
We first examine the overall performance of our modifications on attention mechanisms across 10 million and 100 million word corpora (Figure \ref{fig:blimpbar}). We observe a clear trend in the low-data setting. We see that all models with our modified attention mechanisms (apart from the Primacy and Recency ablations) demonstrate a substantial and statistically significant improvements ($p{\leq}0.001$) in average BLiMP accuracy compared to the baseline model. While the baseline scores approximately 61\% average accuracy on the task, models with fixed attention windows, which impose the most stringent constraints, achieve markedly higher scores of around 68\%. This result highlights the benefit of architectural inductive biases derived from WM principles, particularly when training data is scarce.

However, this advantage diminishes when models are trained on the larger 100M-word dataset. In this higher-data regime, the baseline's performance improves markedly to an accuracy of approximately 71\%, narrowing the performance gap considerably. Notably, models with exponential or primacy-and-recency constraints on attention show no statistically significant difference from the baseline. With a sufficient volume of data, the standard attention mechanism evidently recovers much of the capability required for this task. Despite this, the best-performing constrained models maintain a small but statistically significant edge, indicating that their inductive biases remain beneficial even at a larger scale.

\begin{figure}[!h]
\begin{center}
\includegraphics[width=\linewidth]{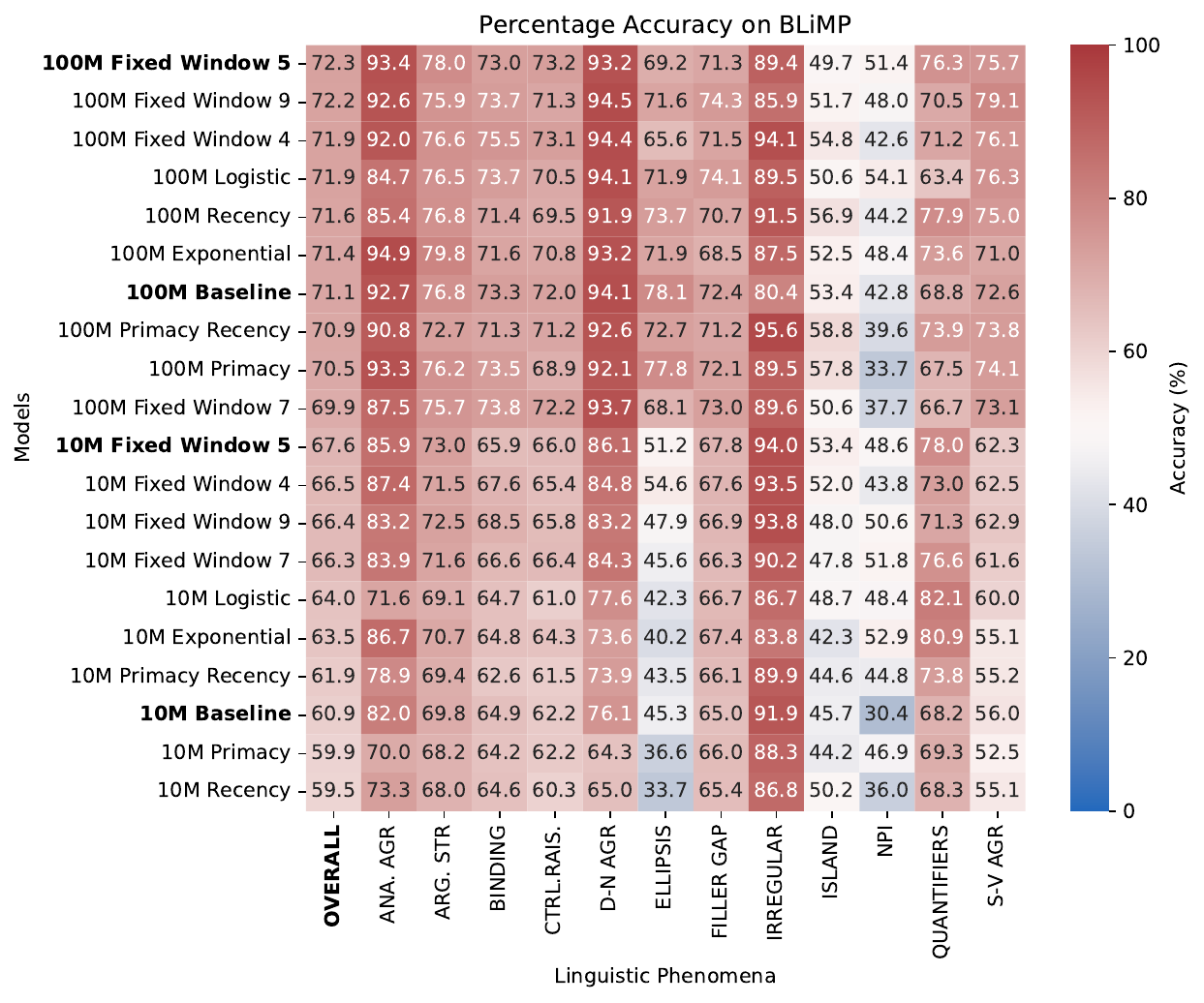}
\end{center}
\caption{Model performance across linguistic phenomena evaluated on the BLiMP task. Models are sorted by Overall score in a descending order; 50\% indicates chance performance. Best performing 10M and 100M models are highlighted along with the corresponding Baselines.}
\label{fig:blimpheatmap}
\end{figure}

\subsection{Performance across linguistic phenomena}
\label{sec:linguisticob}
In order to understand how our modifications to attention influence performance on specific linguistic tasks, we analysed the results across individual BLiMP sub-tasks. In Figure ~\ref{fig:blimpheatmap}, we categorise the different tasks into a set of classes motivated by the analyses in~\citet{warstadt2020blimp}. Consistent with previous findings, our models perform best on phenomena related to morphological agreement, which often rely on local dependencies. Across both 10 million and 100 million data regimes, all models achieve reasonably high accuracy on Determiner-Noun Agreement (which focuses on number agreement, e.g., that chair vs. that chairs), Irregular Forms, and Anaphor Agreement (where the focus is on reflexive pronoun agreement, e.g., girls insulted themselves vs. herself). We note that the fixed-window models, which explicitly enforce locality, excel here. We also find that most models suffer on more abstract syntactic and semantic constraints. Performance on Island Effects (restrictions on syntactic movement) and NPI Licensing (the requirement for words like ever to be in a negative context) is the lowest across the board, often only marginally better than chance. This confirms that these phenomena, which predominantly require sensitivity to complex structural and logical scope, represent a persistent challenge for language models, and our architectural modifications do not offer a simple solution to assist models on these tasks. 

On the other hand, the results for Argument Structure, which governs a verb's ability to appear with certain arguments (e.g., disturbing a person vs. boasting a person), are particularly interesting. Performance is generally boosted by locality constraints across both data regimens. We see that the models that have fixed window size of about 5 and exponentially decaying attention constraints (which tend to promote highly local attention structures) seem to generally perform better\footnote{We also see that the performance on Argument Structure especially is remarkably similar to GPT2-large pretrained model~\cite{warstadt2020blimp}.}.  We also observe that Ellipsis emerges as a phenomenon highly dependent on data scale. At 10M, all models perform exceptionally poorly, with accuracies in the 33-54\% range. However, at 100M words, performance improves dramatically into the 65-78\% range across all models. This sharp increase suggests that the generalisations required to correctly resolve ellipsis are not readily captured with limited data, but indeed become accessible with more exposure.

An important overall finding is that highly constrained, simple models often outperform the less-constrained baseline, particularly on complex syntactic and semantic challenges. Our null hypothesis before experimentation was that heavily constrained models may significantly hamper performance in transformers, especially since a large body of work in contemporary NLP looks into methods for moving away from locality bias \cite[\emph{inter alia}]{tay2020long,zaheer2020big}. We find the performance of small, rigid attention windows on phenomena that are not strictly local to be the most surprising result. For instance, fixed window 5 is broadly one of the best performing models in complex sub categories. Especially in Argument Structure where smaller fixed window models obtain significantly higher performance, where the sentences usually focus on structures which govern a verb's relationship with its arguments. This is noteworthy because these relationships can be complex and are not solely determined by adjacent words, yet this highly local model captures them effectively. Similarly, these models perform well on Binding, which involves the structural relationship between a pronoun and its antecedent. One might expect this to require a wider context, but the small fixed window proves highly effective, outperforming the baseline. On the other hand, leaky models, both Exponential and Logistic bias based models, which allow attention to "leak" across the entire context while prioritising recent information, show interesting trends, however these models are inferior in performance compared to stricter and smaller fixed window based models. 

\paragraph{Primacy/Recency Ablation} We wanted to further understand the efficacy of primacy and recency based formulation and carried out an experiment where we trained models separately with primacy and recency biases. We present these results in Figures \ref{fig:blimpbar} and \ref{fig:blimpheatmap}. We notice, unsurprisingly, that Recency based bias in attention tends to perform better with local structures while Primacy tends to have a slight edge on tasks with dependency structures that require access to longer distances. While both Primacy and Recency generally tend to perform worse in comparison to the baseline model, however, increasing data tends to substantially help the model with Recency based bias. This tends to correlate with fixed window models with strict local attention constraints. Finally, while the model based on Primacy and Recency is not among the best models, however in most cases, it has a tendency to even out the divergence between models trained separately with Primacy or Recency.

\subsection{Analysing the model behaviour with human processing data}
\label{sec:psycoob}
\begin{figure}
    \centering
    \includegraphics[width=\linewidth]{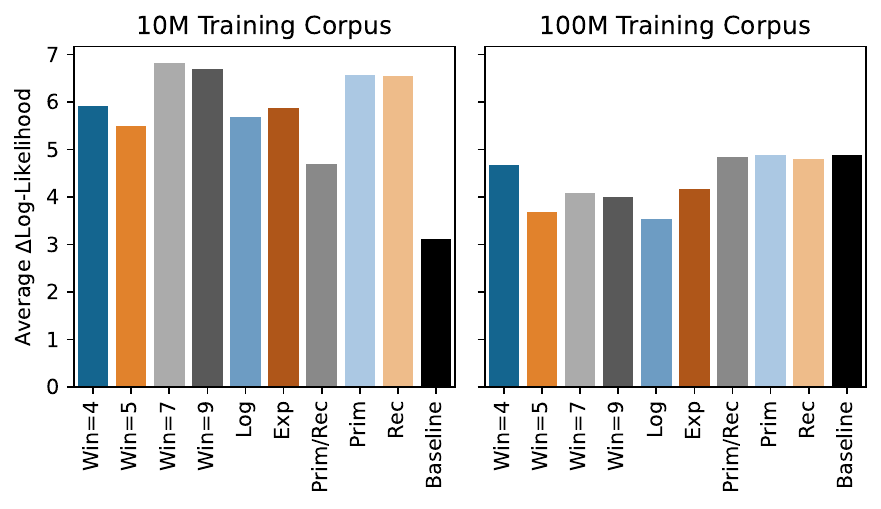}
    \caption{$\Delta$Log-Likelihood averaged across psychometric measures for each model.}
    \label{fig:deltaloglik}
\end{figure}

\begin{figure*}[!htbp]
\begin{center}
\begin{subfigure}[b]{0.49\linewidth}
    \includegraphics[width=\linewidth]{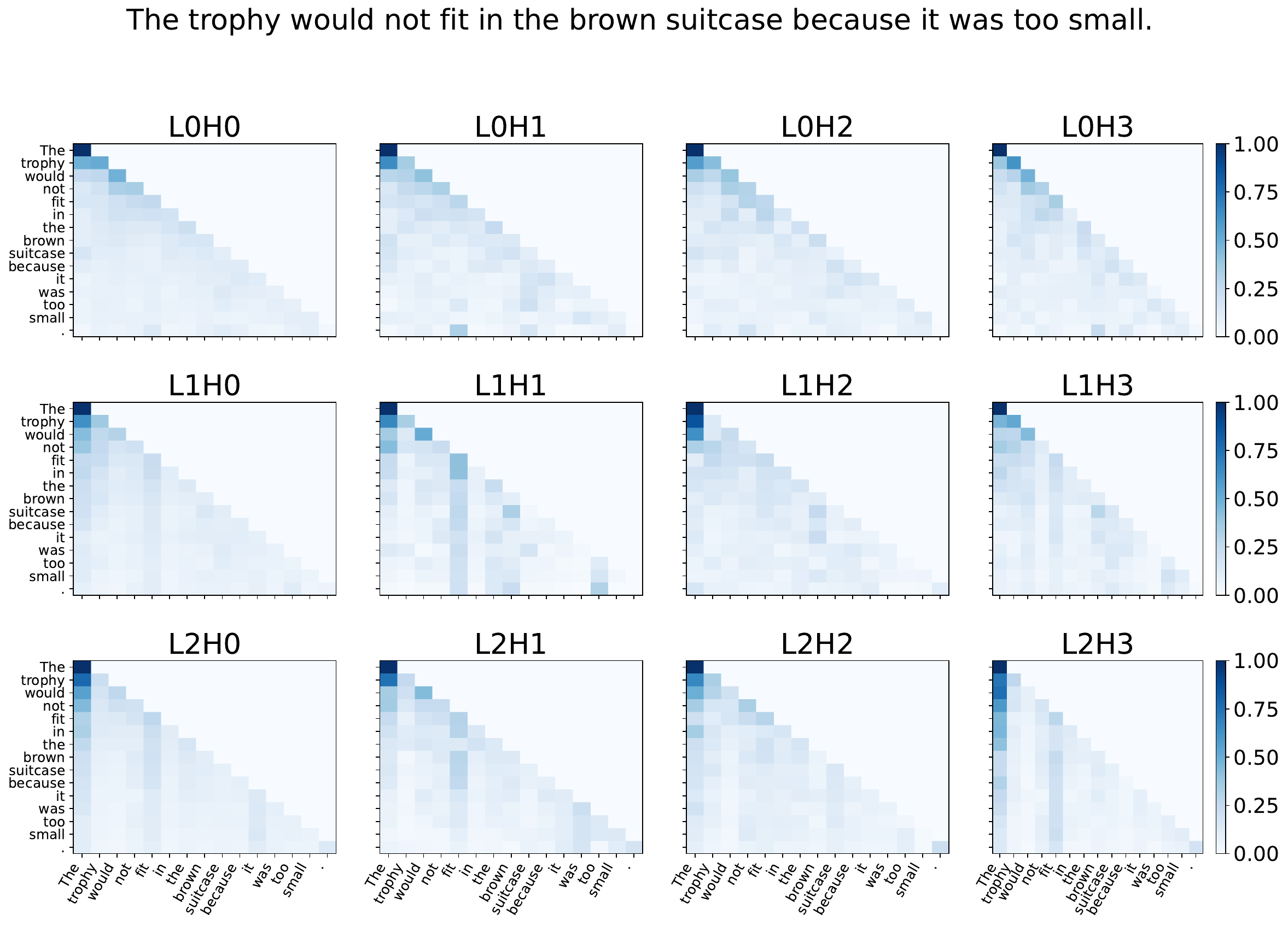}
    \caption{10M Baseline}
    \label{fig:att_a}
\end{subfigure}
\hfill
\begin{subfigure}[b]{0.49\linewidth}
    \includegraphics[width=\linewidth]{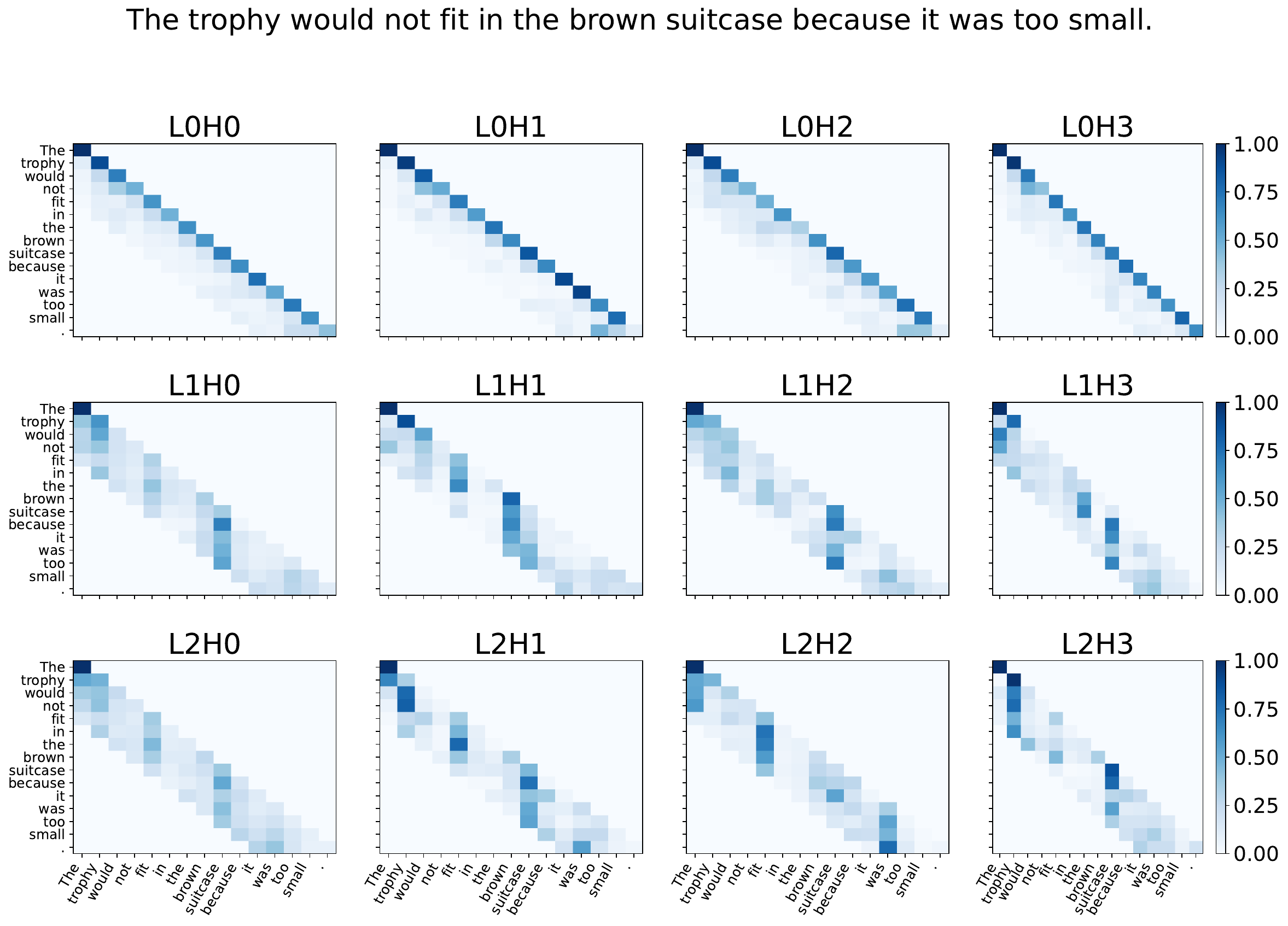}
    \caption{10M Fixed Window 5}
    \label{fig:att_b}
\end{subfigure}
\end{center}
\caption{Attention weights distributions for the sentence "The trophy would not fit in the brown suitcase because it was too small." across individual heads in the first  three layers of the 10M Baseline and Fixed Window 5 models.}
\label{fig:attfig}
\end{figure*}

Would our cognitively inspired attention constraints correlate with human processing data? To answer this we examine the alignment of surprisal and psychometric data using averaged difference of log-likelihood (Figure \ref{fig:deltaloglik}). Here, $\Delta$Log-Likelihood measures the additional explanatory power offered by surprisal values over a base model without surprisal. We then average this across psychometric corpora. 
That is, the difference in log-likelihood ($\Delta$Log-likelihood) represents the improvement in statistical model fit when surprisal is added as a predictor, compared to a base statistical model that includes only covariates like word length and frequency. Therefore, a higher $\Delta$Log-likelihood indicates that a language model's surprisal values are a better predictor of human cognitive effort during sentence comprehension. The `Baseline' bar in the figure refers to the $\Delta$Log-Likelihood achieved by GPT-2 model with unmodified attention trained on corresponding corpora, serving as our primary model for comparison.

We observe a significant trend emerging in the low data regime. Nearly all models with modified attention mechanisms produce surprisal values that are substantially more predictive of human processing data than the unmodified Transformer baseline. The baseline model achieves a relatively low average $\Delta$Log-Likelihood of approximately 3.2, whereas the top-performing models with fixed-window attention (especially 7 and 9), as well as Primacy and Recency biases, achieve scores nearly twice as high. This result suggests that in data-constrained settings, architectural constraints that mimic principles of human WM serve as a powerful and effective inductive bias, guiding the model to learn representations that are more closely aligned with human cognitive processes.

On the other hand, the advantage of these explicit architectural biases diminishes considerably when the models are trained on the larger, 100M-word dataset. The standard baseline model's performance improves, with its average $\Delta$Log-Likelihood increasing marginally. More importantnly, the performance gap between the baseline and the modified models narrows significantly. This pattern suggests that with an order of magnitude more data, the standard self-attention mechanism is better able to approximate the necessary behaviours and produce surprisal values that correlate well with human processing metrics. However, we note here that even for all the constrained attention models and the baseline (with unmodified attention) the $\Delta$Log-likelihood is lower in the 100M setting compared to the 10M setting. A compelling hypothesis for this seemingly counterintuitive trend relates to the divergence between the language modelling objective and the cognitive processes underlying human comprehension. As models are trained on more data, they become increasingly specialised at predicting the statistical patterns of the training corpus. This high degree of specialisation may cause their expectations to diverge from the more generalised predictions that humans make which is also illustrated in recent work \cite[\textit{inter alia}]{oh2023does,de2024cloze,shain2024large}.

\subsection{Understanding attention distribution}
\label{sec:attentionob}

As an initial, exploratory step toward understanding the inductive biases in our models, we examine the internal representations along two complementary
dimensions: the attention patterns within individual heads, and the
syntactic structure recoverable from the contextual embeddings.
Figure \ref{fig:attfig} compares the attention patterns of two
models trained on the 10M corpus, the baseline GPT-2 with
unmodified self-attention and a constrained model with a
fixed-window mechanism (window size=5), on the sentence ``The
trophy would not fit in the brown suitcase because it was too
small'', a classic test case for pronoun resolution. We focus the
figure on early layers, though the pattern is consistent across
higher layers. We observe significant diffences. The fixed-window model
shows an immediate and sharp focus on local context, with heads in
the initial layer already concentrating on the immediately
preceding token. This specialisation sharpens in later layers:
heads \texttt{L2H0} and \texttt{L2H1} come to focus on the core
subject-verb-object structure (`trophy', `fit', `suitcase'), a
pattern reminiscent of the telegraphic speech observed in children
between 18 and 36 months \citep{brown1973first}; \texttt{L2H2}
specialises in verbs (`fit', `was') and \texttt{L2H3} in nouns
(`trophy', `suitcase'). The baseline model mostly shows dissimilar patterns where its heads remain diffuse across layers, attending to
less interpretable combinations of function and content words
without evident functional differentiation. The fixed-window
constraint appears to force a structured division of labour
amongst the attention heads, producing a representation of syntax
that is more explicit than anything that emerges in the
unconstrained model.

Following
\citet{hewitt-manning-2019-structural} and its recent extension by
\citet{someya2025derivational}, we further train a diagnostic linear
projection to reconstruct unlabelled dependency trees from each
model's contextual embeddings, reporting Unlabelled Unrooted
Attachment Score (UUAS) by dependency relation. The fixed-window
model achieves higher UUAS than the baseline across all five
relations tested, with the largest gaps on core grammatical
relations such as \texttt{nsubj} and \texttt{dobj}, which capture
the link between a verb and its arguments. The advantage emerges
early in the network and is most pronounced at intermediate layers.
Full per-layer and per-relation results are in Appendix
\ref{appendix:probing}.

The architectural constraint seems to shape the local attention patterns within heads, producing interpretable functional specialisation. It also tends to alter the global geometry of the representational space such that syntactic dependencies become more linearly recoverable. %

\section{Conclusions}
We investigated the integration of cognitively inspired working
memory constraints into the Transformer architecture, comparing
fixed-width attention windows, exponential and logistic decay
mechanisms, and primacy and recency biases against an unmodified
GPT-2 baseline trained on developmentally plausible corpora of 10M
and 100M words. Across grammatical competence on BLiMP, alignment
with a convergent battery of neural and behavioural processing
measures, and analyses of the models' internal representations, the
results converge on a single picture. Constraints aligned with the
limits of human working memory function as a potent inductive bias,
yielding gains in data efficiency, predictive alignment with human
comprehension, and the development of linguistically interpretable
internal structure. The advantage is most pronounced in the
data-limited regime that most closely resembles the conditions under
which humans acquire language, and the psychometric alignment of all
models, both constrained and unconstrained, degrades as training data
increases, consistent with growing evidence that the
language-modelling objective and human comprehension diverge as
models grow larger. These findings push against the
contemporary trends towards longer contexts and weaker inductive
biases, and suggest that hard cognitive constraints actively scaffold
learning rather than hindering it. %

\newpage
\section*{Limitations}

We believe several bounds on the present work deserve explicit acknowledgement. Our investigation is deliberately confined to the developmentally
plausible regimes of 10M and 100M words, and we make no claim about
what happens at the trillion-token scale of contemporary frontier
models. The trajectory of psychometric alignment we observe between
10M and 100M, in which alignment degrades rather than improves with
more data, is at least consistent with the possibility that the
specialised structure our constrained models acquire reflects an
inductive bias absent from the unconstrained architecture rather
than a pattern recoverable from data alone, but our experiments don't currently dont provide strong experimental evidence that this is the case with all dominant models. Our implementation of working
memory is also a deliberate simplification. Working memory is not
a rigid, position-indexed buffer. It is perhaps a dynamic, content-addressable system shaped by similarity-based interference among held items \citep{van2003distinguishing} and by linguistic experience itself \citep{schwering2020verbal, macdonald2016speak}. In our experiments, we isolate one dimension, namely the locality and decay of accessible context, and
the development of attention mechanisms reflecting interference and
content-addressable retrieval is the natural next step. Our
empirical evaluation is also restricted to English, and the
locality constraints that benefit our models here may interact
quite differently with languages whose dependencies are mediated by say morphological case, by free word order, or by head-final
configurations.

A more interesting class of limitations comes from the results
themselves. While constrained models perform well across most BLiMP
categories, their performance on Island Effects remains only
marginally above chance, mirroring the unconstrained baseline. This
raises the genuine possibility that locality constraints, useful as
they are for the acquisition of local syntactic dependencies, may
impede the acquisition of phenomena that require sensitivity to
global syntactic structure. Whether cognitive constraints have
natural analogues that operate over abstract \emph{structural} rather than linear distance is, to us, among the more interesting directions this work opens up. The most fundamental bound, however, is one our study shares with nearly all contemporary computational
psycholinguistics. Human language is acquired, produced, and
comprehended in densely multimodal contexts where gesture, prosody,
gaze, and pragmatic interaction contribute centrally to meaning
\citep[see][for a broader perspective]{holler2019multimodal}. Working memory in everyday language
use is integrated across these multiplicities of modalities and record of experiences. Our work, like the overwhelming majority of
language modelling research, operates within a unimodal text-only
paradigm, and the cognitive plausibility achievable within that
paradigm is correspondingly bounded. Whether the architectural constraints we study have analogues that would prove useful in genuinely multimodal language models is perhaps the most interesting question this work leaves open.

\section*{Acknowledgements}
This work was supported in part by the Alan Turing Institute under Fundamental Research (Project No.\ PP00029). 
\bibliography{custom,anthology}
\clearpage

\appendix

\section{Architectural Parameters and Training Configuration}
\label{appendix:architecture}

\subsection{Base Architecture}

All models in this study use the GPT-2 small architecture as a
base, with the following parameters: 12 transformer decoder
layers, 768-dimensional token embeddings and hidden states, 12
parallel attention heads per layer, and a maximum context window
of 1024 tokens. The combination yields approximately 124 million
trainable parameters per model. The results are averaged over 3 runs.

\subsection{Per-Mechanism Parameter Justifications}

\paragraph{Fixed Window Attention.} We evaluate window sizes
$W \in \{4, 5, 7, 9\}$. The choice of $W = 4$ is motivated by
\citet{cowan2001magical}, who proposed an upper bound of
approximately four chunks on the contents of focal attention in
short-term memory under controlled conditions. The choices of
$W \in \{5, 7, 9\}$ are derived from \citet{miller1956magical},
whose ``magical number seven, plus or minus two'' represents the
classical estimate of the number of items maintainable in
immediate memory. In the present setting, we treat each token as
a fundamental information chunk, acknowledging that this is a
simplification of human chunking, which operates over
linguistically meaningful units of variable size.

\paragraph{Exponential Decay Attention.} We adopt
$\lambda = 82.86$ and $\alpha = 0.37$, the values reported as
optimal by \citet{de-varda-marelli-2024-locally} from a grid
search using GPT-2 small on the Provo corpus
\citep{luke2018provo}. We retain these values without
re-tuning in order to enable direct methodological comparison
with the post-hoc application reported in that work.

\paragraph{Logistic Decay Attention.} We use $k = 0.4$ as the
steepness parameter and $m = 12.0$ as the midpoint distance.
These values establish a profile where tokens within
approximately five positions retain high attention weight before
the decay accelerates.

\paragraph{Primacy and Recency Attention.} The mechanism includes
two learnable scalar parameters, $w_{\text{primacy}}$ and
$w_{\text{recency}}$, which control the relative contribution of
primacy and recency biases. Both are initialised at 0.5 at the
start of training. %

\subsection{Training Configuration}

All models are trained from scratch using the AdamW optimiser
with a learning rate of $5 \times 10^{-5}$, batch size of 64,
and weight decay of 0.01. Training proceeds for 5 epochs with
gradient clipping at a maximum norm of 1.0. These hyperparameters
follow the empirical setups commonly used in the BabyLM Challenge
\citep{warstadt-etal-2023-findings} and are kept consistent
across all model variants to ensure fair comparison. Tokenisation
uses the standard GPT-2 byte-pair encoding, with separate
tokenisers trained on the 10M and 100M corpora respectively to
match the data scale of each setting.

\begin{figure*}[!h]
\begin{center}
\includegraphics[width=\linewidth]{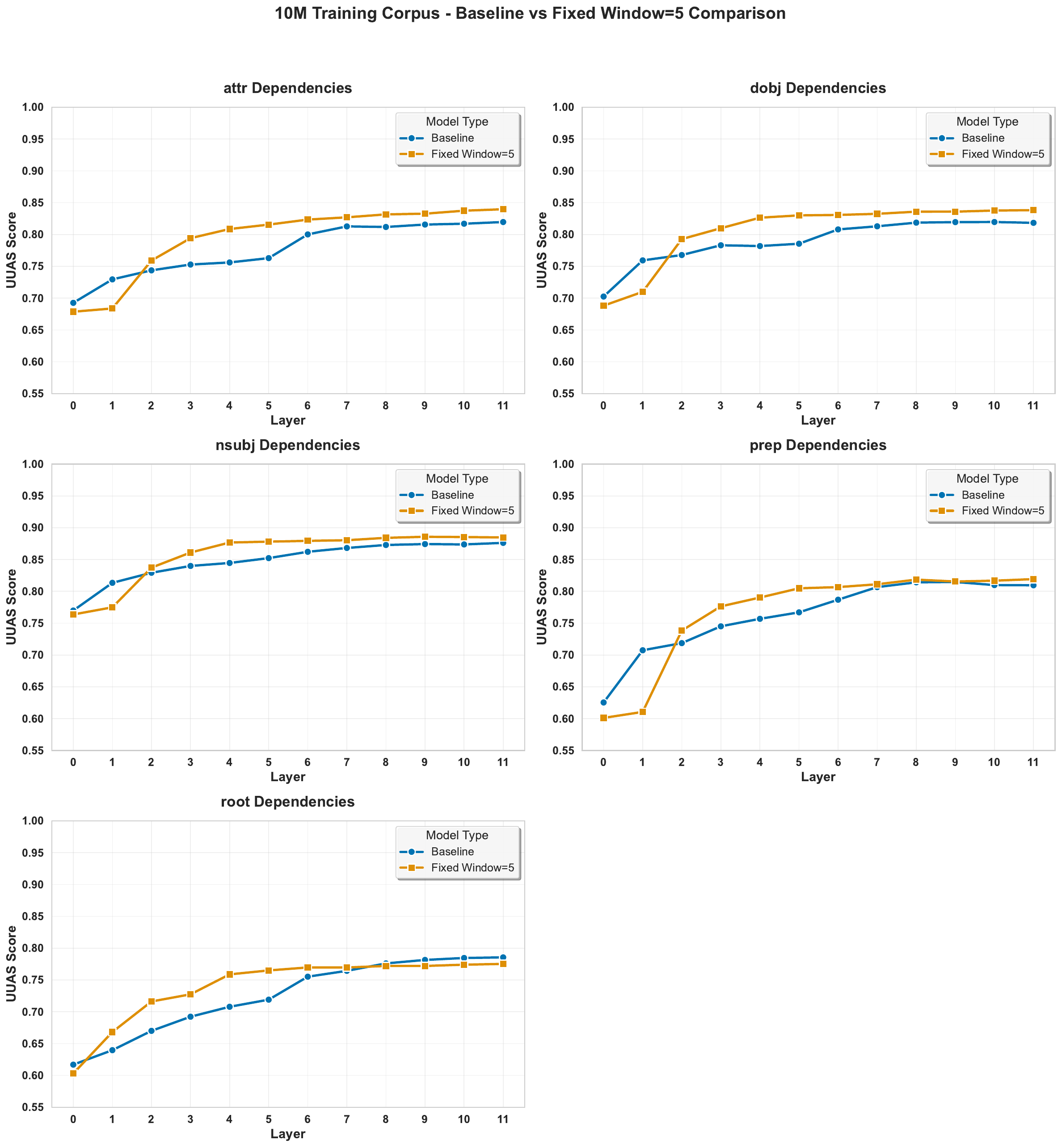}
\end{center}
\caption{UUAS Score comparison between the Baseline and Fixed Window 5 models trained on 10M words.}
\label{fig:uuas}
\end{figure*}

\section{Structural Probing Analysis}
\label{appendix:probing}

Our structural probing analysis applies the derivational probing
framework of \citet{someya2025derivational}, which extends the
structural probe of \citet{hewitt-manning-2019-structural} by
making explicit the layer-wise derivation of syntactic structures
in neural language models. The method trains a diagnostic linear
projection $B \in \mathbb{R}^{k \times d}$ over a model's
contextual embeddings $h_i \in \mathbb{R}^d$, such that the
squared $L_2$ distance between projected embeddings approximates
the tree distance between the corresponding tokens in the
syntactic dependency parse:

\begin{equation}
d_B(h_i, h_j)^2 = (B(h_i - h_j))^T (B(h_i - h_j))
\end{equation}

The probe is trained to minimise the difference between $d_B$ and
the true parse-tree distance $d_T$ over a training set of parsed
sentences. From the resulting distance matrix at evaluation time,
we recover the unlabelled syntactic dependency structure using a
minimum spanning tree algorithm, and we report the Unlabeled
Unrooted Attachment Score (UUAS) by relation type. UUAS measures
the proportion of edges in the recovered tree that match an
edge in the gold parse, ignoring direction and dependency labels.

We probe representations at every layer of the model (layers 0--11
for GPT-2 small) and report results for the five most frequent
dependency relations in our evaluation corpus: \texttt{nsubj}
(nominal subject), \texttt{dobj} (direct object), \texttt{prep}
(prepositional modifier), \texttt{attr} (attribute), and
\texttt{root} (sentence root). Probes are trained on
gold-parsed sentences from the English Web Treebank, with
embeddings extracted from each model in inference mode (no
gradient flow back to the language model). Probe dimensionality
is set to $k = 64$, optimised with Adam at a learning rate of
$10^{-3}$ for 30 epochs, with early stopping on validation loss.
We evaluate on a held-out test partition.

We believe that this structural probing complements behavioural evaluation in a specific way. BLiMP measures whether a model assigns higher probability to grammatically acceptable sentences than to
unacceptable ones, but does not directly tell us whether the
model has internalised the structural relations that underlie
the contrast. A model could in principle perform well on
agreement minimal pairs by tracking surface co-occurrence
statistics rather than by representing the binding relation
between subject and verb. In a way, structural probing addresses this gap by asking whether the geometry of a model's representations
encodes the grammatical structure of its inputs in a recoverable
form. A higher UUAS indicates that syntactic dependencies are
linearly decodable from the embedding space, which we take as
evidence that the model has acquired a more explicit internal
representation of structure rather than an implicit, behaviourally
adequate proxy.

\subsection{Results by Dependency Relation}

Figure \ref{fig:uuas} reports UUAS for the baseline and
fixed-window models trained on the 10M corpus, plotted as a
function of layer depth. Several patterns are worth noting.
First, the fixed-window model achieves consistently higher UUAS
than the baseline across all five relations, with the gap
emerging in the early-to-middle layers and persisting across the
remainder of the network. The advantage is most pronounced for
\texttt{nsubj} and \texttt{dobj}, where the fixed-window model
reaches UUAS values approximately 0.05 above the baseline at
several intermediate layers. These two relations capture the
core argument structure of clauses, which is precisely the
domain in which the fixed-window model also outperforms the
baseline most dramatically on BLiMP. The convergence of the
behavioural and probing evidence is notable, and suggests that
the BLiMP gains observed for the fixed-window model are
underwritten by genuinely better structural representations
rather than by surface-level heuristics.

Second, both models converge to similar UUAS scores at the
deepest layers for several relations, particularly \texttt{prep}
and \texttt{root}. This is consistent with the general finding
in structural-probing literature that the latest layers of
language models tend to be optimised for the prediction
objective rather than for representational explicitness, and
that intermediate layers often carry more directly recoverable
structural information \citep{hewitt-manning-2019-structural,
tenney-etal-2019-bert}. The fixed-window model's advantage is
therefore most visible precisely where syntactic structure is
most explicitly available, namely the early-to-middle layers.

Third, the fixed-window model develops its structural
representations earlier in the network. For \texttt{nsubj},
\texttt{dobj}, and \texttt{prep}, the fixed-window model's
UUAS at layer 2 is comparable to the baseline's UUAS at layer
4 or 5. This earlier emergence of structure is consistent with
our attention-distribution analysis (Section~\ref{sec:attentionob}), which suggests that
the fixed-window model's heads specialise for syntactic functions
already in the early layers. Together, these results suggest
that the architectural constraint pushes the model to discover
syntactic abstractions sooner in the processing pipeline,
freeing later layers for higher-order operations.

\subsection{Relation to the Attention-Head Analysis}

The probing results provide a quantitative complement to the
qualitative attention-head analysis in Section~\ref{sec:attentionob}. Where the
attention-head visualisations show that individual heads in the
fixed-window model specialise for identifiable linguistic
functions, the probing results show that this specialisation
translates into representational geometry in which syntactic
relations are linearly recoverable. The two analyses converge
on the same underlying claim that the fixed-window constraint
shapes both the local attention patterns within heads and the
global geometry of the resulting representational space.

We however caution that structural probing remains an indirect measure.
A high UUAS shows that syntactic structure is recoverable from
the embeddings, but does not establish that the model uses this
structure in any computationally meaningful way during
prediction. The convergence between probing results and
behavioural performance on BLiMP is suggestive. We treat the probing analysis as one of several
converging lines of evidence rather than as a standalone proof
of structural representation.

\section{Detailed BLiMP Analysis by Phenomenon}
\label{appendix:blimp_detail}

\begin{figure*}[!ht]
\begin{center}
\resizebox{0.8\linewidth}{!}{
\includegraphics{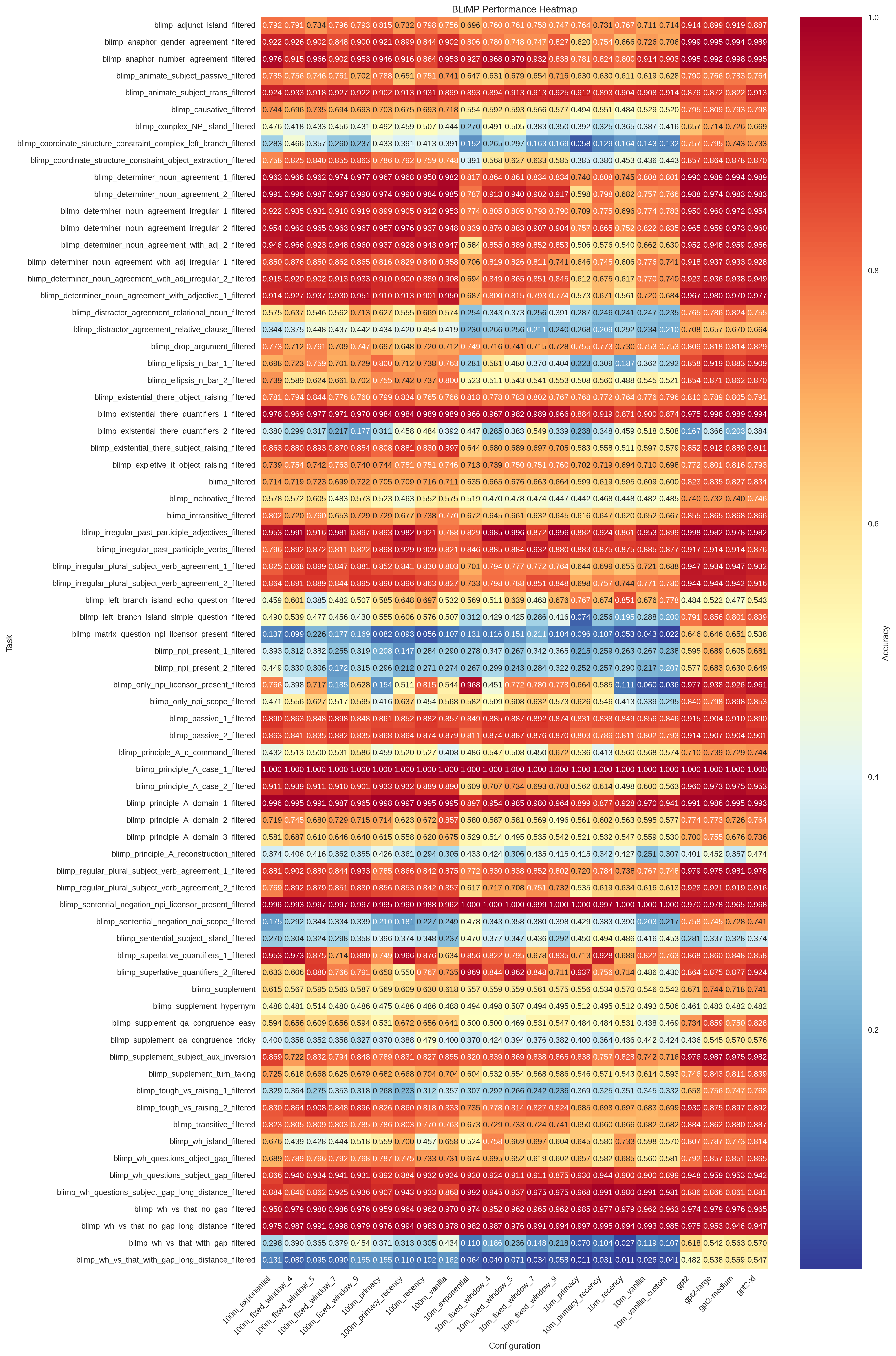}}
\end{center}
\caption{Model performance on BLiMP across individual tasks and models.}
\label{fig:blimp_heatmap2}
\end{figure*}

The main text reports BLiMP performance aggregated by linguistic
category (Section \ref{sec:linguisticob}). Here we provide a finer-grained analysis at the
level of individual sub-tasks, drawing on the per-phenomenon
heatmap shown in Figure \ref{fig:blimp_heatmap2}. The
analysis surfaces several patterns that do not emerge clearly at
the category level and that bear on the interpretation of where
architectural constraints help, where they do not, and why.

The largest constrained-versus-baseline gaps appear on phenomena
that combine two properties: a syntactically structured
relationship between elements that are not always linearly
adjacent, and a heavy dependence on training data for the
acquisition of that relationship. Argument structure phenomena
highlight this pattern. Sub-tasks such as
\texttt{animate\_subject\_passive},
\texttt{animate\_subject\_trans}, and
\texttt{causative} test whether the model has internalised the
selectional restrictions verbs impose on their arguments. The
fixed-window models with $W=5$ achieve accuracies on these
sub-tasks at the 10M scale that often match or exceed the
corresponding 100M baseline, suggesting that locality is doing
genuine work in scaffolding verb-argument acquisition rather
than merely capturing surface co-occurrence. This pattern is
consistent with the argument structure of English clauses being
predominantly local, with the verb's arguments typically
appearing within a few tokens of the verb itself.

Binding phenomena show a parallel pattern. Sub-tasks such as
\texttt{principle\_A\_c\_command} and
\texttt{principle\_A\_case} test sensitivity to the structural
constraints governing the relationship between reflexive pronouns
and their antecedents. Constrained models, particularly fixed
windows of size 5, perform competitively on these tasks despite
the fact that binding relations are not always strictly local.
This is theoretically interesting: the principle of binding is a
structural constraint over c-command relations rather than over
linear distance, and one might expect that models with restricted
linear access would struggle. The empirical pattern suggests
either that the relevant binding relations in BLiMP minimal pairs
typically fall within the model's window, or that the
representational structure encouraged by the locality constraint
generalises usefully to capturing structural rather than purely
linear dependencies. Distinguishing these two possibilities
empirically would require a controlled analysis on minimal pairs
stratified by the linear distance between binder and bindee, which
we leave to future work.

Two categories show consistently weak performance across all
models, including the constrained variants: Island Effects and
NPI Licensing. The performance ceiling on these categories is
substantially below the model's overall accuracy, and the
constrained models offer no clear advantage over the baseline.
We take this as informative rather than disappointing.

Island Effects test the model's sensitivity to constraints on
syntactic movement, such as the wh-island and complex NP-island
constraints. These constraints are not reducible to linear
distance or to local agreement; they require sensitivity to
abstract syntactic configurations that hold over arbitrarily
extensive structural domains. A model whose architectural
constraint is defined in terms of linear distance, as ours are,
has no principled mechanism for representing such configurations.
The persistence of poor performance across all our locality-based
mechanisms therefore aligns with theoretical expectation: linear
locality is the wrong abstraction for capturing island
constraints. A more interesting open question, raised in the
Limitations section, is whether cognitive constraints have
natural analogues that operate over abstract structural distance,
which might capture island phenomena where linear constraints
cannot.

NPI Licensing tests the model's sensitivity to the requirement
that negative polarity items such as \emph{ever} or \emph{any}
appear in the scope of a licensing operator such as negation or
a question. The constraint is logical and scope-based rather
than structural in the syntactic sense, and the relevant
licensing operator may be arbitrarily distant from the NPI in
linear terms. As with island constraints, a locality-based
constraint is unlikely to offer a principled solution here, and
the empirical results bear this out. Performance on
\texttt{npi\_present\_1} and related sub-tasks remains close to
chance for most models at both training scales.

The heatmap also reveals systematic differences among the
fixed-window variants themselves. Window size 5 emerges as the
most consistent performer across phenomena, achieving the highest
overall accuracy at 10M and remaining competitive at 100M.
Windows of size 7 and 9 perform comparably to size 5 on most
sub-tasks but show occasional degradation on phenomena requiring
particularly local sensitivity, such as some morphological
agreement tasks. Window size 4 exhibits the opposite pattern,
performing well on local phenomena but showing larger
degradations on sub-tasks involving moderately non-local
dependencies. The convergence on $W=5$ as the best-performing
choice is itself empirically interesting and may reflect
something about the typical phrasal length over which English
syntactic dependencies operate, although we emphasise that this
observation is suggestive rather than conclusive without
controlled cross-linguistic replication.

The exponential and logistic decay variants show a different
profile. They perform competitively with the fixed-window models
on phenomena requiring strict locality but underperform on tasks
where information from slightly more distant tokens is
informative. We attribute this to the soft nature of the decay
constraint: while the bias favours nearby tokens, attention to
more distant tokens is not strictly prevented, and the model
appears to retain some capacity to attend non-locally when the
training signal calls for it. Whether this represents a genuine
weakness of soft decay relative to hard windowing, or whether it
reflects inadequate hyperparameter tuning of the decay
mechanisms, is a question we cannot fully resolve here.

\section{AI Assistant Use}
We made use of generative AI tools to assist in the drafting of the manuscript and the refinement of the research code. These tools were used only for linguistic clarity and code scaffolding (VS Code). All technical content and final interpretations were rigorously reviewed and verified by us.

\end{document}